\documentclass{article}
\usepackage{iclr2022_conference,times}

\usepackage{amsmath,amsfonts,bm}

\def\eqref#1{Equation~\ref{#1}}

\def\1{\bm{1}}

\DeclareMathAlphabet{\mathsfit}{\encodingdefault}{\sfdefault}{m}{sl}
\SetMathAlphabet{\mathsfit}{bold}{\encodingdefault}{\sfdefault}{bx}{n}

\DeclareMathOperator*{\argmin}{arg\,min}

\iclrfinalcopy
\usepackage[utf8]{inputenc} 
\usepackage[T1]{fontenc}    
\usepackage{hyperref}       
\usepackage{url}            
\usepackage{booktabs}       
\usepackage{nicefrac}       
\usepackage{microtype}      
\usepackage{stmaryrd}
\usepackage{lipsum}
\usepackage{amsfonts}
\usepackage{graphicx}
\usepackage{epstopdf}
\usepackage{subcaption}
\ifpdf
  \DeclareGraphicsExtensions{.eps,.pdf,.png,.jpg}
\else
  \DeclareGraphicsExtensions{.eps}
\fi

\usepackage{enumitem}
\setlist[enumerate]{leftmargin=.5in}
\setlist[itemize]{leftmargin=.5in}

\title{Low-Rank+Sparse Tensor Compression for Neural Networks}

\author{Cole Hawkins\thanks{Work completed during an internship at Facebook}\\ University of California, Santa Barbara \AND Haichuan Yang, Meng Li, Liangzhen Lai, Vikas Chandra\\
Facebook Inc.}

\usepackage[ruled,vlined]{algorithm2e}
\usepackage{algorithmic}
\usepackage{amsmath}
\usepackage{amsopn}
\usepackage{amssymb}
\usepackage{placeins}
\usepackage{xcolor}

\DeclareMathAlphabet\mathbfcal{OMS}{cmsy}{b}{n}

\newtheorem{definition}{Definition}
\newcommand{\ten}[1]{\mathbfcal{#1}} 
\newcommand{\mat}[1]{\mathbf{#1}}
\newcommand{\fm}[1]{\mat{U}^{(#1)}}
\newcommand{\ft}[1]{\ten{G}^{(#1)}} 

\begin{document}

\maketitle

\begin{abstract}
Low-rank tensor compression has been proposed as a promising approach to reduce the memory and compute requirements of neural networks for their deployment on edge devices. Tensor compression reduces the number of parameters required to represent a neural network weight by assuming network weights possess a coarse higher-order structure. This coarse structure assumption has been applied to compress large neural networks such as VGG and ResNet. However modern state-of-the-art neural networks for computer vision tasks (i.e. MobileNet, EfficientNet) already assume a coarse factorized structure through depthwise separable convolutions, making pure tensor decomposition a less attractive approach. We propose to combine low-rank tensor decomposition with sparse pruning in order to take advantage of both coarse and fine structure for compression. We compress weights in SOTA architectures (MobileNetv3, EfficientNet, Vision Transformer) and compare this approach to sparse pruning and tensor decomposition alone.

\end{abstract}

\section{Introduction}\label{sec:intro}

Using machine learning to enable multiple applications on edge devices requires multiple task-specific persistent models \citep{yang2020co}. These models are used for tasks ranging from computer vision \citep{howard2019searching} to automatic speech recognition \citep{wang2021noisy}. The trend towards multiple applications and multiple models is constrained by the fact that off-chip memory reads incur high latency and power costs \citep{sze2017hardware}. Therefore, in this paper we target memory cost reduction. In this area low-rank tensor compression is a popular approach \citep{garipov2016ultimate,novikov2015tensorizing,lebedev2014speeding} that can achieve orders of magnitude compression, but can lead to significant accuracy loss. 

Low-rank tensor compression has achieved impressive headline compression numbers \citep{garipov2016ultimate} and is suitable for on-device acceleration \citep{zhang2021fpga,deng2019tie} due to its reliance on standard dense linear algebra operations. However it is often applied to massively overparameterized architectures such as VGG or ResNet \citep{novikov2015tensorizing,hawkins2019bayesian}. Recent work benchmarking sparse pruning \citep{blalock2020state} raises the issue that compression techniques applied to overparameterized architectures may not reach the compression/accuracy Pareto frontier of SOTA compressed networks such as EfficientNet. Further, tensor compression enforces low-rank factorized structure on neural network weights. This factorized structure is already exploited by SOTA computer vision backbones via depth-wise separable convolutions. This motivates us to consider the following question: how can low-rank tensor compression contribute to SOTA computer vision architectures? In this paper we investigate whether low-rank tensor compression can be combined with sparse pruning to capture complementary coarse and fine structures respectively and outperform either sparse pruning or low-rank tensor factorization alone.

Specifically, we investigate two forms of low-rank plus sparse decomposition of neural network weights. First we consider additive structure in which the neural network weights can be decomposed as the sum of a low-rank component and a sparse component. Second we consider a low-rank or sparse structure in which the entries of neural network weights are taken from either a sparse pruned weight or a low-rank weight.

\subsection{Related Work}

Existing work has studied the combination of low-rank matrices and sparse representations to compress VGG-style models \citep{yu2017compressing} and ResNets \citep{gui2019model}. These methods apply additive low-rank plus sparse compression to the weights of neural networks and can achieve compression results that outperform sparse compression or low-rank compression alone. More recent work suggests that low-rank structure is a viable compression method for the EfficientNet model class \citep{gu2021towards} on CIFAR-10/100 computer vision tasks. An open question is whether low-rank matrix or tensor structure remains for efficient modern networks trained on ImageNet.

\subsection{Contributions}

Our main contribution is our study of low-rank structure remaining in recent SOTA efficient neural networks for computer vision tasks. We observe that trends in architecture design \citep{howard2017mobilenets,sandler2018mobilenetv2,howard2019searching} have removed low-rank structure from such networks by building in factorized weights. Therefore the direct application of low-rank matrix and tensor methods is challenging. To the best of our knowledge, our work is the first to consider the combination of low-rank tensor compression with sparse pruning. Further, our work is the first to study low-rank+sparse compression of weights for SOTA architectures that rely on efficient depthwise separable convolutions.

\begin{figure*}
    \centering
    \includegraphics[width=\textwidth]{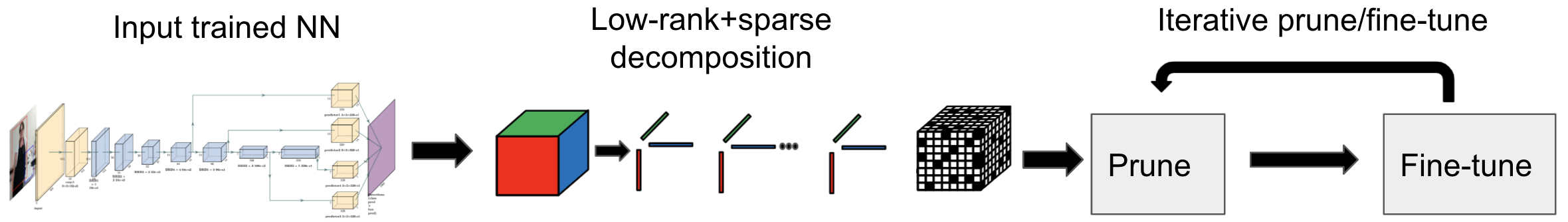}
    \caption{High-level proposed algorithm flow.}
    \label{fig: algo flow}
\end{figure*}

\section{Background}

\subsection{Tensor Formats}

\begin{figure}[t]
\centering
  \begin{subfigure}[t]{0.35\textwidth}
  \centering
    \includegraphics[width=.35\textwidth]{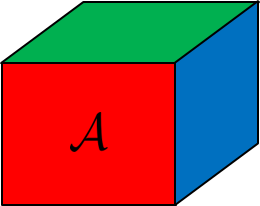}
    \caption{\label{fig: full visual}}
  \end{subfigure}
  \hspace{10pt}
\begin{subfigure}[t]{0.35\textwidth}
  \centering
    \includegraphics[width=1.0\textwidth]{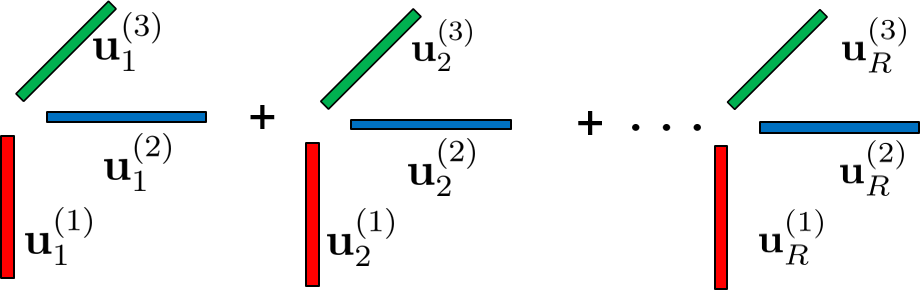}
    \caption{\label{fig: cp visual description}}
  \end{subfigure}
  \\

     \begin{subfigure}[t]{0.35\textwidth}
  \centering
    \includegraphics[width=.9\textwidth]{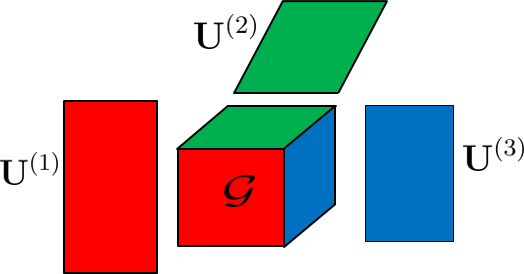}
    \caption{\label{fig: tucker visual description}}
  \end{subfigure}
    \hspace{10pt}
     \begin{subfigure}[t]{0.35\textwidth}
  \centering
    \includegraphics[width=1.0\textwidth]{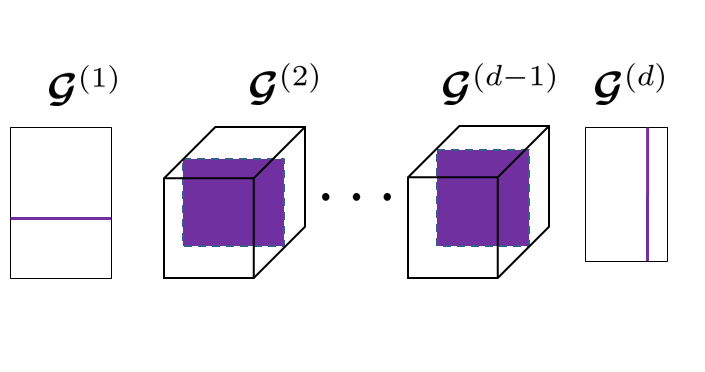}
    \caption{\label{fig: tt visual description}}
  \end{subfigure}
  \caption{Taken from \cite{hawkins2020towards}. (a): An order-3 tensor, (b) and (c): representations in CP and Tucker formats respectively, where low-rank factors are color-coded to indicate the corresponding modes. (d): TT representation of an order-$d$ tensor, where the purple lines and squares indicate $\ten{G}^{(n)}(:,i_n,:)$, which is the $i_n$-th slice of the TT core $\ten{G}^{(n)}$ obtained by fixing its second index. \label{fig: visual formats}}
\end{figure}

We follow the notation and exposition of \citep{hawkins2020towards} to describe the low-rank tensor formats that we employ. This paper uses lower-case letters (e.g., $a$) to denote scalars, bold lowercase letters (e.g., $\mat{a}$) to represent vectors, bold uppercase letters (e.g., $\mat{A}$) to represent matrices, and bold calligraphic letters (e.g., $\ten{A}$) to denote tensors. A tensor is a generalization of a matrix, or a multi-way data array. An order-$d$ tensor is a $d$-way data array $\ten{A}\in \mathbb{R}^{I_1 \times I_2 \times \dots \times I_d}$, where $I_n$ is the size of mode $n$. The $(i_1, i_2, \cdots, i_d)$-th element of $\ten{A}$ is denoted as $a_{i_1i_2 \cdots i_d}$. An order-$3$ tensor is shown in Fig.~\ref{fig: visual formats} (a).

The number of entries in a tensor with mode size $n$ and $d$ modes is $n^d$. This exponential dependence on the tensor order is often referred to as the ``curse of dimensionality" and can lead to high costs in both computing and storage. Fortunately, many practical tensors have a low-rank structure, and this property can be exploited to reduce such costs. We will use four tensor decomposition formats (CP, Tucker, and TT/TTM) to reduce the parameters of neural networks.

First, we introduce a necessary decomposition to describe the CP format.
\begin{definition}
A $d$-way tensor $\ten{A} \in \mathbb{R}^{I_1\times \cdots \times I_d}$ is rank-1 if it can be written as a single outer product of $d$ vectors
\begin{equation}
\ten{A} =\mat{u}^{(1)} \circ \dots \circ \mat{u}^{(d)}, \; {\text{with}} \; \mat{u}^{(n)} \in\mathbb{R}^{I_n} \; \text{for}\; n=1,\cdots, d. \nonumber
\end{equation}
\end{definition}
Each element of $\ten{A}$ is $a_{i_1 i_2 \cdots i_d}=\prod \limits_{n=1}^d u_{i_n}^{(n)}$,
where $u_{i_n}^{(n)}$ is the $i_n$-th element of the vector $\mat{u}^{(n)}$.

\begin{definition}\label{Def: CP}
The CP factorization~\citep{carroll1970analysis,harshman1994parafac} expresses tensor $\ten{A}$ as the sum of multiple rank-1 tensors:
\begin{equation}
\ten{A} = \sum_{j=1}^R \mat{u}_j^{(1)} \circ \mat{u}_j^{(2)} \dots \circ \mat{u}_j^{(d)}.
\end{equation}
Here $\circ$ denotes an outer product operator. The minimal integer $R$ that ensures the equality is called the {\bf CP rank} of $\ten{A}$. To simplify notation we collect the rank-1 terms of the $n$-th mode into a factor matrix $ \fm{n} \in \mathbb{R}^{I_n \times R}$ with $\fm{n}(:,j) = \mat{u}_{j}^{(n)}$. A rank-$R$ CP factorization can be described with $d$ factor matrices $\{ \mat{U}^{(n)}\}_{n=1}^d$ using $R\sum_n I_n$ parameters.
\end{definition}

Next we introduce a prerequisite definition for the low-rank Tucker format.
\begin{definition}
The  mode-$n$ product of a tensor $\ten{A}\in\mathbb{R}^{I_1\times\dots \times I_{n} \times \dots\times I_d}$ with a matrix $\mat{U}\in\mathbb{R}^{J\times I_n}$ is 
\begin{equation}
    \label{eq: mode n product}
    \begin{split}
    \ten{B} &= \ten{A}\times_n \mat{U} \Longleftrightarrow
    b_{i_1\dots i_{n-1} j i_{n+1} \dots i_d}=\sum_{i_n=1}^{I_n} a_{i_1  \dots i_d}  u_{j i_n}  .
    \end{split}
\end{equation}
\end{definition}
The result is still a $d$-dimensional tensor $\ten{B}$, but the mode-$n$ size becomes $J$. In the special case $J=1$, the $n$-th mode diminishes and $\ten{B}$ becomes an order-$d-1$ tensor. The Tucker format is defined by a series of mode$-n$ products between a small core tensor and several factor matrices.

\begin{definition}\label{Def: Tucker}
The Tucker factorization~\citep{tucker1966some} expresses a $d$-way tensor $\ten{A}$ as a series of mode-$n$ products:
\begin{equation}
\label{eq:Tucker}
\ten{A} = \ten{G}\times_1 \fm{1}\times_2 \dots \times_d \fm{d}.
\end{equation}
Here $\ten{G}\in\mathbb{R}^{R_1\times \dots\times R_d}$ is a small core tensor, and $\fm{n}\in\mathbb{R}^{I_n\times {R}_n}$ is a factor matrix for the $n$-th mode. The {\bf Tucker rank} is the tuple $(R_1,\dots,R_d)$. A Tucker factorization with ranks $R_n=R$ requires $R^d+R\sum_n I_n$ parameters.
 \end{definition}
The product structure permits more expressive cross-dimension interactions than the CP model but may not result in large parameter reductions due to the exponential dependence on the tensor order $d$.

Finally we introduce two formats, Tensor-Train and Tensor-Train Matrix. Like the Tucker format, these formats permit more expressive interactions across dimensions. Unlike the Tucker format these formats remove the exponential dependence on the tensor order $d$.
\begin{definition}
\label{def: tensor train}
The tensor-train (TT) factorization~\citep{oseledets2011tensor} expresses a $d$-way tensor $\ten{A}$ as a collection of matrix products:
\begin{equation}
\label{eq:TT}
a_{i_1 i_2 \dots i_d} = \ft{1}(:,i_1,:)\ft{2}(:,i_2,:)\dots \ft{d}(:,i_d,:).
\end{equation}
Each  TT-core $\ft{n}\in \mathbb{R}^{R_{n-1}\times I_n \times R_{n}}$ is an order-$3$ tensor. The tuple $(R_0,R_1,\dots,R_d)$ is the {\bf TT-rank} and $R_0=R_d=1$.
\end{definition}
The TT format uses $\sum_n R_{n-1}I_nR_{n}$ parameters in total and leads to more expressive interactions than the CP format.

The TT-format is well-suited to tensors, but is equivalent to low-rank matrix format when the order is $d=2$. We can extend the TT-format to matrices by reshaping along each mode. Let $\mat{A}\in\mathbb{R}^{I\times J}$ be a matrix. We assume that $I$ and $J$ can be factored as follows:
\begin{equation}
\label{eq: TTM dimension factorization}
    I=\prod_{n=1}^d I_n,J=\prod_{n=1}^d J_n.
\end{equation}
We can reshape $\mat{A}$ into a tensor $\ten{A}$ with dimensions $I_1\times\dots\times I_d\times J_1\times\dots\times J_d$, such that the $(i,j)$-th element of $\mat{A}$ uniquely corresonds to the $(i_1, i_2, \cdots, i_d, j_1, j_2, \cdots, j_d)$-th element of $\ten{A}$. The TT decomposition can extended to compress the resulting order-$2d$ tensor as follows.

\begin{definition}
\label{def: tensor train matrix}
The tensor-train matrix (TTM) factorization expresses an order-$2d$ tensor $\ten{A}$ as $d$ matrix products: 
\begin{equation}
\label{eq: ttm factorization}
a_{i_1\dots i_d j_1 \dots j_d} = \ft{1}(:,i_1,j_1,:)\ft{2}(:,i_2,j_2,:)\dots \ft{d}(:,i_d,j_d,:).
\end{equation}
Each TT-core $\ft{n}\in \mathbb{R}^{R_{n-1}\times I_n \times J_n \times R_{n}}$ is an order $4$ tensor. The tuple $(R_0,R_1,R_2,\dots,R_d)$ is the {\bf TT-rank} and as before $R_0=R_d=1$. This TTM factorization requires $\sum_n R_{n-1}I_n J_nR_{n}$ parameters to represent $\ten{A}$. \end{definition}

We provide a visual representation of the CP, Tucker, and TT formats in Fig.~\ref{fig: visual formats} (b) -- (d).

\subsection{Folding to high-order tensors.} In the previous section we described how to adapt the TT format to the matrix case by factorizing the dimensions of a matrix and viewing the matrix as a tensor. Our goal is to apply tensor decompositions to the weights of many architectures that include $1\times 1$ convolutions (MobileNet, EfficientNet) or fully-connected layers (Vision Transformer). These matrix-format weights are well-suited to the TTM decomposition, but not the CP and Tucker formats which collapse to low-rank matrix format. To apply the CP and Tucker decompositions to these matrix weights we can perform a reshape operation. A weight matrix $\mat{W}\in\mathbb{R}^{I\times J}$ can be folded into an order-$d$ tensor $\ten{A}\in\mathbb{R}^{I_1\times\dots\times I_d}$ where $IJ=\prod_n I_n$. We can also fold $\mat{W}$ to an order-$2d$ tensor $\ten{A} \in \mathbb{R}^ {I_1 \times \cdots \times I_d \times J_1 \times \cdots \times J_d }$ such that $w_{ij}=a_{i_1\cdots i_d j_1 \cdots j_d}$. While a convolution filter is already a tensor, we can reshape it to a higher-order tensor with reduced mode sizes. Reshaping imposes additional higher-order and hierarchical tensor structure.

\subsection{Sparse Pruning}
We prune a tensor weight $\ten{S}$ by applying a binary tensor mask $\ten{M}$ to $\ten{S}$ using the entry-wise product
\[
(\ten{M}\odot \ten{S})_{i_1\dots i_d}=m_{i_1\dots i_d}\times s_{i_1\dots i_d}.
\]
We focus on global magnitude pruning \cite{lecun1990optimal} throughout this work and use a cubic pruning schedule taken from \cite{zhu2017prune}. At each step we mask the bottom $100\times(s_t-s_{t-1})$ percent of the non-masked weights in $\ten{S}$ by setting their corresponding entries in $\ten{M}$ to zero where 
\begin{equation}
    \label{eq: pruning}
        s_t = s_f\left(\frac{t}{n}\right)^3
\end{equation}
and $s_f$ is the desired final sparsity ratio, $t$ is the current step, and $n$ is the total number of steps. This process gradually increases the sparsity in $\ten{M}$ until the target sparsity $s_f$ is reached.

\subsection{Distillation}

The goal of knowledge distillation is to transfer the knowledge of an expensive teacher model to a cheaper student model \citep{hinton2015distilling}. In this setting ``expensive" refers to the memory cost of storing the neural network weights. The expensive teacher model $f_t$ is a neural network pretrained on ImageNet. The student model $f_s$ is the compressed teacher model obtained through sparse pruning and/or tensor compression. Given a training example with input $\mat{x}$ and label $\mat{y}$ the knowledge distillation loss is 
\begin{equation}
    \label{eq: KD loss}
    \mathcal{L}\left(\mat{x},\mat{y}\right)=\alpha l(\mat{x},\mat{y})+(1-\alpha)T^2 KL\left(f_s(\mat{x})/T||f_t(\mat{x})/T\right)
\end{equation}
where $l$ is any standard loss function and $KL$ is the KL-divergence. In our setting $l$ is the cross-entropy loss. The KL term in the loss function penalizes the student output distribution when it differs from the teacher output distribution. Both outputs are temperature scaled by $T$. We set $\alpha=0.9$ and $T=3$ which is a standard choice \citep{hinton2015distilling,cho2019efficacy}. We note that another reasonable loss function is fine tuning on the training data only (set $\alpha=1$) which we do not explore in this work.


\section{Method: Low-Rank + Sparse Tensor Decomposition}

At a high level, our algorithm takes a trained neural network $f_t$ as input and then outputs a low-rank plus sparse tensor-compressed neural network $f_s$ by {\bf Phase 1:} low-rank plus sparse tensor decomposition and {\bf Phase 2:} iterative pruning and fine-tuning. This is outlined in Figure \ref{fig: algo flow}. The technical description is given in Algorithm \ref{alg: outer}.

\begin{algorithm}
\caption{Low-Rank + Sparse Compression and Fine-Tuning}
\begin{algorithmic}
\label{alg: outer}
\REQUIRE Trained NN $f_t$, tensor ranks $\{\mat{r}_i\}_{i=1}^L$, sparsity target $s_f$, number of epochs $e$
\FOR{$i$ in $[1,2,3,\dots]$}
\STATE Compress layer $i$ weight $\ten{A}_i$ with target rank $\mat{r}_i$ using \eqref{eq: tensor residual minimization} or \eqref{eq: tensor masking residual minimization}.
\ENDFOR
\FOR{$t$ in $[1,2,3,\dots,e]$}
\STATE Global magnitude pruning on sparse residuals $\{\ten{S}_l\}_{i=1}^L$ with target sparsity $s_t$ from \eqref{eq: pruning}.
\STATE Train for one epoch using loss function $\mathcal{L}$ from \eqref{eq: KD loss}.
\ENDFOR
\end{algorithmic}
\end{algorithm}

\subsection{Reconstruction}

We consider two approaches to represent the tensor $\ten{A}$. We assume that the reconstruction $\ten{A}=h(\ten{L},\ten{S})$ combines the low-rank and sparse components and is performed before any forward propagation. Depending on the specific form of $h$, which will be specified below, direct contraction of the layer activation with $\ten{S}$ and $\ten{L}$ respectively may be more efficient.

\paragraph{Additive:} Given weight tensor $\ten{A}$ we reconstruct 
\begin{equation}
    \label{eq: additive reconstruction}
    \ten{A} = \ten{L}+\ten{M}\odot\ten{S}
\end{equation}
where $\ten{L}$ is a low-rank tensor whose weights are maintained in low-rank format and $\ten{S}$ is a sparse residual whose weights will be pruned. The binary mask $\ten{M}$ controls the sparsity of $\ten{S}$. After decomposition, each forward pass of the neural network will reconstruct the weight $\ten{A}\leftarrow \ten{L}+\ten{M}\odot\ten{S}$, permitting backpropagation directly to the low-rank tensor factors comprising $\ten{L}$ and the sparse unpruned entries of $\ten{S}$. For example, when the low-rank tensor $\ten{L}$ is in CP format we store the factor matrices $\{\mat{U}^{(i)}\}$ and not the full tensor $\ten{L}$. 

\paragraph{Masking:} We identify a drawback of the reconstruction specified in \eqref{eq: additive reconstruction}. First, representational capacity is wasted when the low-rank tensor $\ten{L}$ and the sparse tensor $\ten{S}$ both contribute to an entry in $\ten{A}$. Each nonzero entry in $\ten{S}$ has no constraint, and so the given entry could be represented by $\ten{S}$ alone. This would permit the low-rank factors of $\ten{L}$ to fit other entries in $\ten{A}$ where the sparse pruned $\ten{S}$ is zero and cannot contribute. Therefore we propose an alternate ``masking" reconstruction method in which we reconstruct $\ten{A}$ as 
\begin{equation}
    \label{eq: masking reconstruction}
    \ten{A} = (1-\ten{M})\odot\ten{L}+\ten{M}\odot\ten{S}.
\end{equation}
Under this reconstruction each entry of $\ten{A}$ is represented by either $\ten{L}$ or $\ten{S}$, but not both.

\subsection{Initialization} 

The sparse weight $\ten{S}$ will be iteratively pruned. Therefore our goal is to capture any possible coarse-grained structure in $\ten{A}$ using $\ten{L}$, and retain fine structure in $\ten{S}$ that will be well-suited to sparse pruning. We describe two methods of decomposition. The tensor $\ten{A}$ may be reshaped before performing factorization. 

\paragraph{Residual:}
A simple approach is to obtain $\ten{L}$ by minimizing the residual to $\ten{A}$. We focus on the Tensor-Train format for notational convenience and use
\begin{equation}
    \label{eq: reconstruction}
    \ten{L}=\llbracket \ten{G}^{(1)},\dots,\ten{G}^{(d)}\rrbracket
\end{equation}
to represent the reconstruction of the low-rank TT-format tensor $\ten{L}$ from its factors $\{\ten{G}^{(i)}\}_{i=1}^d$.
\begin{equation}
    \label{eq: tensor residual minimization}
    \begin{split}
    \ten{G}^{(1)},\dots,\ten{G}^{(d)} &= \argmin_{\ten{G}^{(1)},\dots,\ten{G}^{(d)}}\left\|\llbracket \ten{G}^{(1)},\dots,\ten{G}^{(d)}\rrbracket-\ten{A}\right\|_2\\
    \ten{L} &= \llbracket\ten{G}^{(1)},\dots,\ten{G}^{(d)}\rrbracket\\
    \ten{S} &= \ten{A}-\ten{L}.
    \end{split}
\end{equation}  
We obtain an approximate solution to the minimization problem through standard decomposition methods implemented in Tensorly \citep{kossaifi2016tensorly}. For the TT and Tucker formats the decomposition to obtain the factors of $\ten{L}$ requires $d-1$ tensor unfoldings and singular value decompositions. The CP format uses alternating least squares (ALS) iterations. In general, the decomposition time for the networks we consider requires less than five minutes and is inconsequential compared to the resources used during the fine-tuning phase which takes approximately 12hrs.

The decomposition given in \eqref{eq: tensor residual minimization} provides an exact reconstruction of the the original tensor $\ten{A}$ before the sparse component is pruned at the first iteration. We note that this naive decomposition has one clear weakness: it does not take into account the fact that many entries of $\ten{S}$ will be pruned. A natural approach would be to consider robust PCA or other decomposition methods that explicitly target a low-rank plus sparse decomposition. We attempted this approach but did not observe any clear gains. We use the residual initialization approach with the additive approach. 

\paragraph{Masking:} In order to directly encourage the low-rank factors to capture only the low-magnitude entries (which will be pruned) we consider an alternate decomposition. We mask the top-$K\%$ of entries using the operator $P_\Omega$:
\begin{equation}
    \label{eq: masking}
    P_{\Omega}(\ten{A})_{i_1\dots i_d} = 
    \begin{cases}
    0 \hspace{0.3in}\text{if }\hspace{0.1in} a_{ijk}\in TopK(\ten{A})\\
    a_{ijk} \hspace{0.1in}\text{  otherwise}
    \end{cases}
\end{equation}
Then we obtain our low-rank decomposition $\left(\ten{S},\ten{L}\right)$ by minimizing the reconstruction loss and considering only the bottom $K\%$ of entries in $\ten{A}$.
\begin{equation}
    \label{eq: tensor masking residual minimization}
    \begin{split}
    \ten{G}^{(1)},\dots,\ten{G}^{(d)} &= \argmin_{\ten{G}^{(1)},\dots,\ten{G}^{(d)}}\left\|P_{\Omega}\left(\llbracket \ten{G}^{(1)},\dots,\ten{G}^{(d)}\rrbracket-\ten{A}\right)\right\|_2\\
    \ten{L} &= \llbracket\ten{G}^{(1)},\dots,\ten{G}^{(d)}\rrbracket\\
    \ten{S} &= \ten{A}-\ten{L}.
    \end{split}
\end{equation}
This decomposition differs from the one obtained in \eqref{eq: tensor residual minimization} in that the tensor decomposition does not fit any of the high-magnitude entries which will remain unpruned. We use this initialization approach with the ``masking" reconstruction so that the initialization ignores high-magnitude entries that will remain un-pruned.

\section{Experiments}\label{sec: experiments}

In this section we report results obtained on ImageNet-1K. All results are obtained by training on a machine with 32 CPU cores, 8 Tesla V100 GPUs, and 128GB memory. We train each method for 50 epochs using a batch size of 256 and distillation settings $\alpha=0.9,T=3$. Each run takes approximately 12hrs. All reported ImageNet accuracy metrics are validation accuracy. In all experiments we set the tensor rank so that the low-rank factors required to represent $\ten{L}$ use $10\%$ as many elements as are in $\ten{A}$.

\subsection{Reconstruction and Initialization}

\begin{figure}
     \centering
     \begin{subfigure}[t]{0.48\textwidth}
         \centering
         \includegraphics[width=\textwidth]{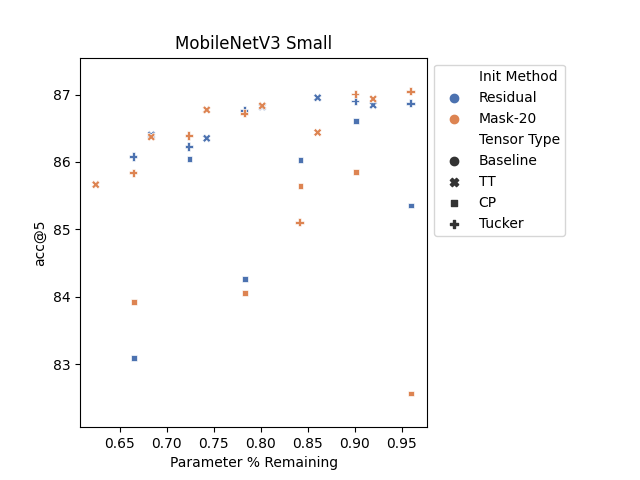}
         \caption{}
         \label{fig: mobilenetv3 small init}
     \end{subfigure}
     \hfill
     \begin{subfigure}[t]{0.48\textwidth}
         \centering
         \includegraphics[width=\textwidth]{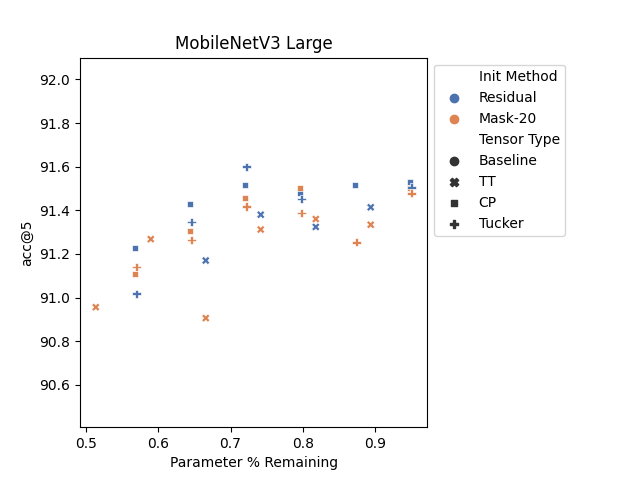}
         \caption{}
         \label{fig: mobilenetv3 large init}
     \end{subfigure}
    \caption{Comparison of initialization strategies for low-rank+sparse tensor decomposition for the (a) MobileNet-v3-Small and (b) MobileNet-v3-Large architectures. There are no major distinctions between initialization strategies.\label{fig: initialization}}
\end{figure}

First we compare two variants of our method: additive reconstruction with residual initialization and masking reconstruction with masking initialization. For this experiment, and all other MobileNet experiments, we use SGD with momentum $0.9$ base learning rate $0.1$ and a decay of $0.7$ applied every $5$ epochs. The decay rate was selected using a small grid search over $\{0.5,0.7,0.9\}$ and selecting the best ImageNet-1K validation accuracy for the sparse pruned MobileNet-v3-Large baseline. In Figure \ref{fig: initialization} we compare the results of our residual initialization strategy with our masking initialization strategy. We observe no clear gains from either strategy. Based on these results we use the simpler residual initialization strategy with additive reconstruction for the remainder of our experiments.

\subsection{MobileNetv3 Results}
\begin{figure}
     \centering
     \begin{subfigure}[t]{0.49\textwidth}
         \centering
         \includegraphics[height=2.0in]{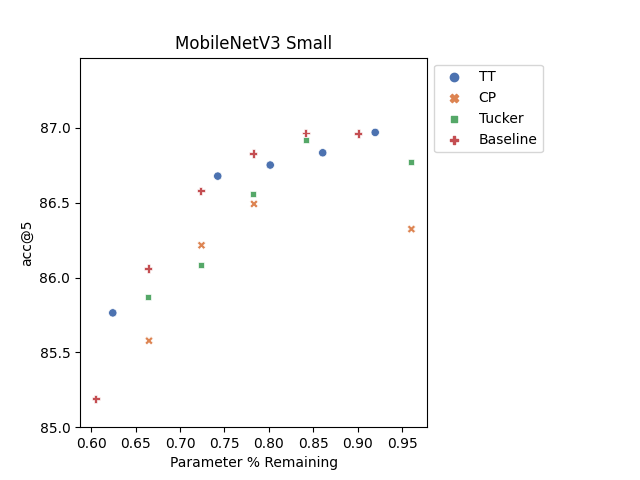}
         \caption{}
         \label{fig: mobilenetv3 small results}
     \end{subfigure}
     \hfill
     \begin{subfigure}[t]{0.49\textwidth}
         \centering
         \includegraphics[height=2.0in]{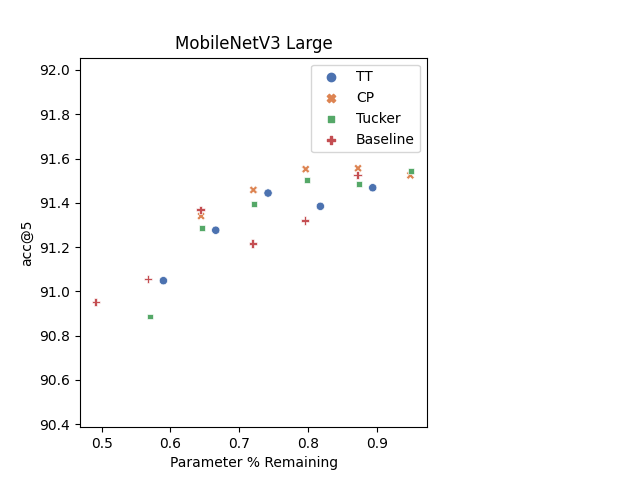}
         \caption{}
         \label{fig: mobilenetv3 large results}
     \end{subfigure}
     \centering
     \begin{subfigure}[t]{0.49\textwidth}
         \centering
         \includegraphics[height=2.0in]{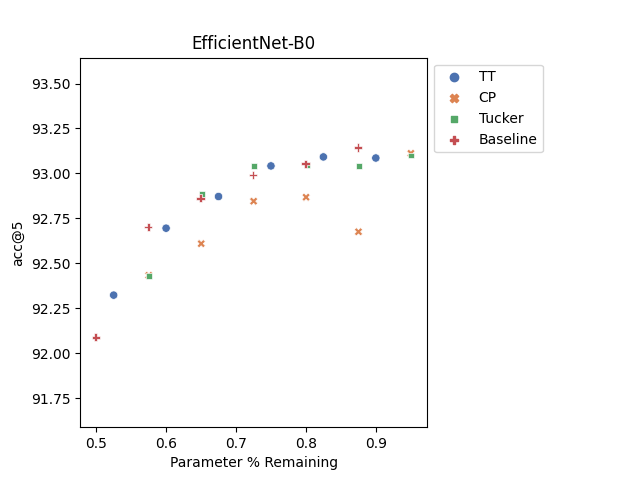}
         \caption{}
         \label{fig: efficientnet b0 results}
     \end{subfigure}
     \hfill
     \begin{subfigure}[t]{0.49\textwidth}
         \centering
         \includegraphics[height=2.0in]{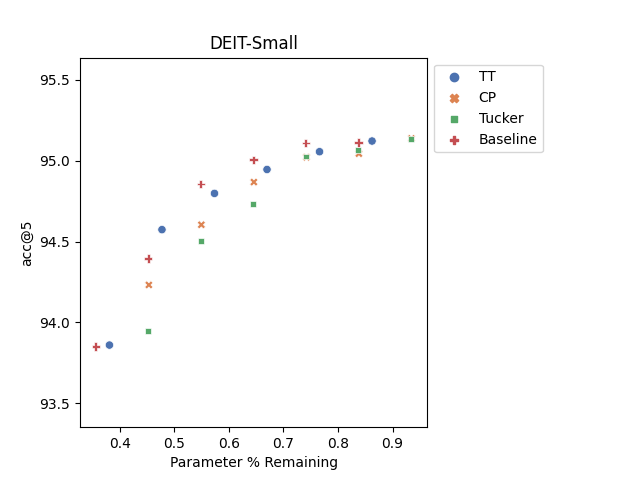}
         \caption{}
         \label{fig: deit small results}
     \end{subfigure}
\caption{ImageNet top-5 accuracy comparison between our LR+S method with CP/TT/Tucker tensor format and the sparse pruning baseline. Results for (a) Mobilenet-v3-Small (b) MobileNet-v3-Large (c) EfficientNet-B0 (d) DEIT-Small.\label{fig: all results}}
\end{figure}

In Figures \ref{fig: mobilenetv3 small results} and \ref{fig: mobilenetv3 large results} we plot a comparison between our method with varying tensor formats (CP,Tucker,TT) and the sparse pruning baseline on MobileNet-v3-Small and MobileNet-v3-Large respectively. We observe that the combination of low-rank and sparse representations is competitive with the sparse pruning baseline. However we do not observe clear performance advantages.

\subsection{Other Models: EfficientNet and DEIT}

To test if our conclusions hold across architectural families we test on EfficientNet-B0 \citep{tan2019efficientnet}. We use the same optimizer settings as the MobileNet experiments, and report our results in Figure ~\ref{fig: efficientnet b0 results}. We also test our results on DEIT \citep{touvron2021training} which is a highly accurate Vision Transformer \citep{dosovitskiy2020image}. Specifically, we the the DEIT-Small model with resolution 224 and patch size 16. We train with the Adam optimizer with learning rate $1e-4$. In Figures \ref{fig: efficientnet b0 results} and \ref{fig: deit small results} we observe similar results as those from the MobileNet family of models. Low-rank plus sparse decomposition is competitive with sparse pruning across models, but does not significantly outperform sparse pruning.

\subsection{Effect on Model Weights}

\begin{figure}
     \centering
     \begin{subfigure}[b]{0.49\textwidth}
         \centering
         \includegraphics[width=\textwidth]{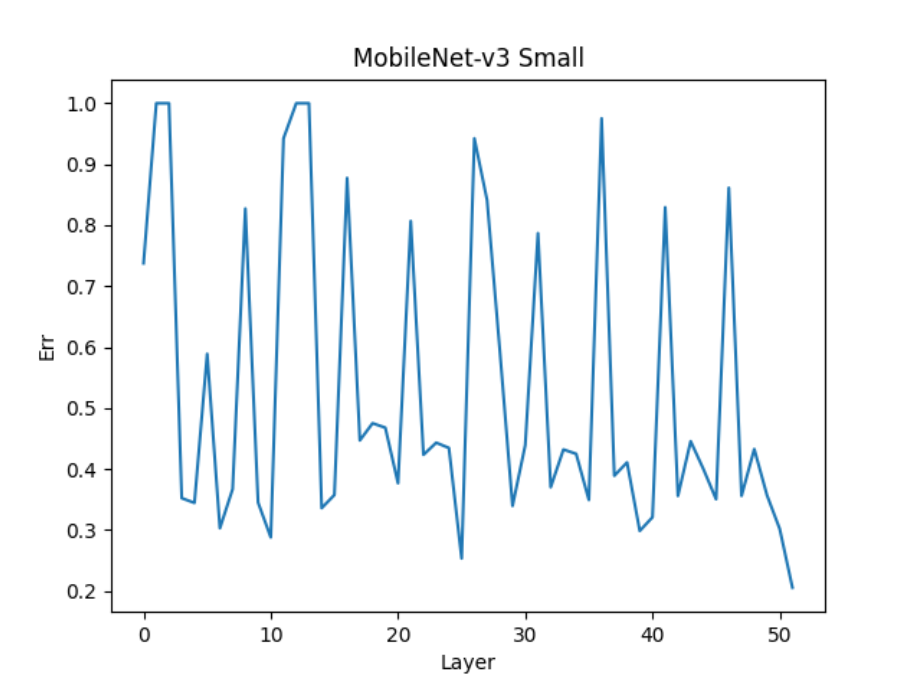}
         \caption{}
         \label{fig: mobilenetv3 layerwise residuals}
     \end{subfigure}
     \hfill
     \begin{subfigure}[b]{0.49\textwidth}
         \centering
         \includegraphics[width=0.9\textwidth]{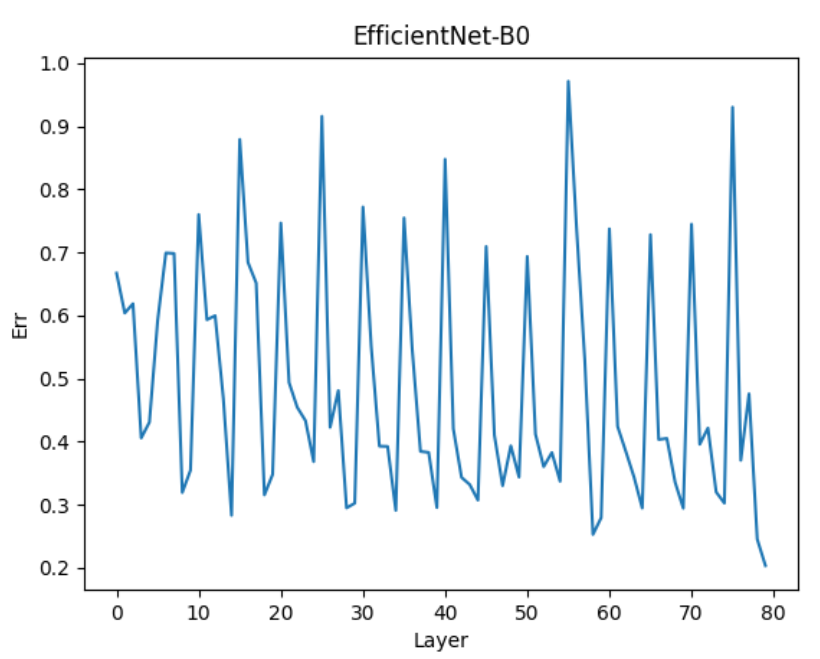}
         \label{fig: efficientnet b0 layerwise residuals}
         \caption{}
     \end{subfigure}
\caption{Layer-wise relative error of the low-rank tensor-train factorization for each layer of (a) Mobilenet-v3-Small and (b) EfficientNet-B0. Other tensor types are similar. Layers with high relative error are the pointwise convolutions.\label{fig: layerwise residuals}}
\end{figure}

In this section we investigate the effects of our compression methods on the model weights. We highlight a key challenge: many pre-trained models do not possess weights that are close to low-rank. In particular, models with a large number of FC layers or 1x1 convolutions are not ``approximately low-rank" even after reshaping from matrix to tensor format. In Figure \ref{fig: layerwise residuals} we plot the layer-wise relative errors of the low-rank decomposition, computed as
\begin{equation}
    \label{eq: relative error}
    Err = \frac{\|\ten{A}-\ten{L}\|_2}{\|\ten{A}\|_2}
\end{equation}
We plot the result of CP-format decomposition applied to the MobilNetv3-small and EfficientNet-B0 models. We observe that many layers have a high relative error, but also that several layers stand out in that the low-rank tensor component fits the original tensor $\ten{A}$ almost exactly.

\begin{figure}[ht]
     \centering
     \begin{subfigure}[b]{0.45\textwidth}
         \centering
         \includegraphics[width=\textwidth]{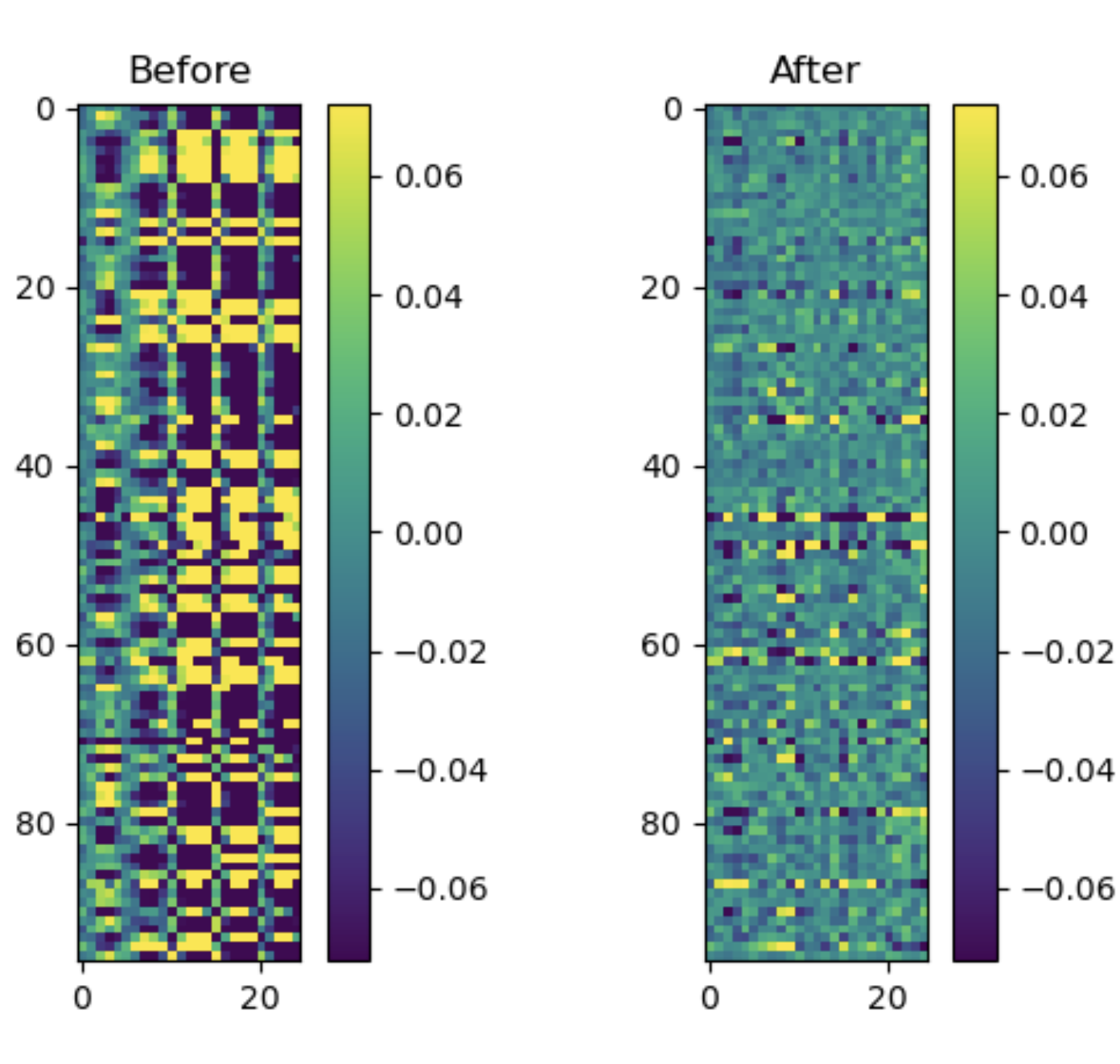}
         \caption{MobileNetv3-Small spatial convolution before and after removing low-rank tensor factors.}
         \label{fig: mobilenetv3_spatial_visual}
     \end{subfigure}
     \hfill
     \begin{subfigure}[b]{0.45\textwidth}
         \centering
         \includegraphics[width=\textwidth]{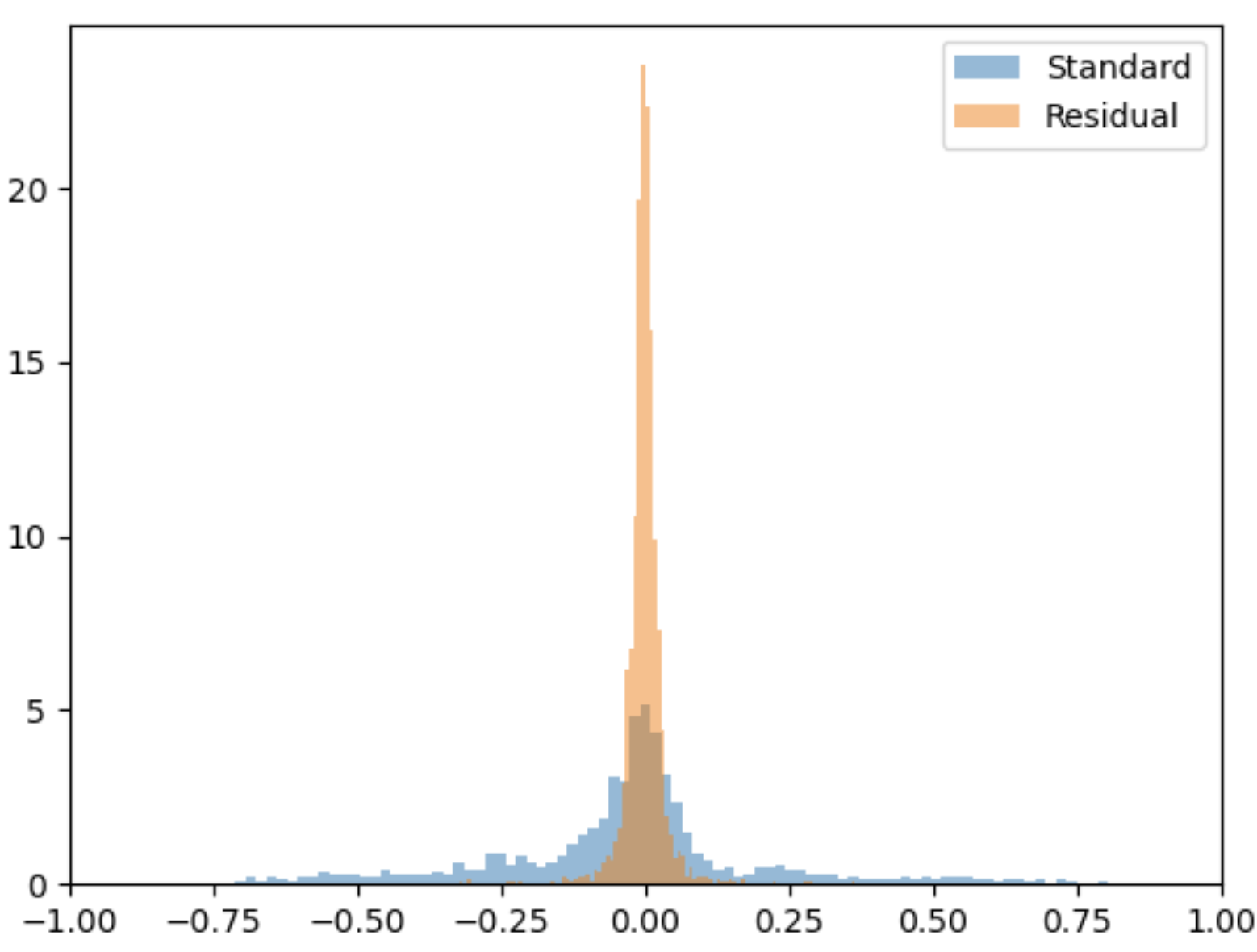}
         \caption{Effect of tensor factorization on MobileNetv3-Small spatial convolution weights.}
         \label{fig: mobilenetv3_spatial_hist}
     \end{subfigure}
     \hfill
     \begin{subfigure}[b]{0.45\textwidth}
         \centering
         \includegraphics[width=\textwidth]{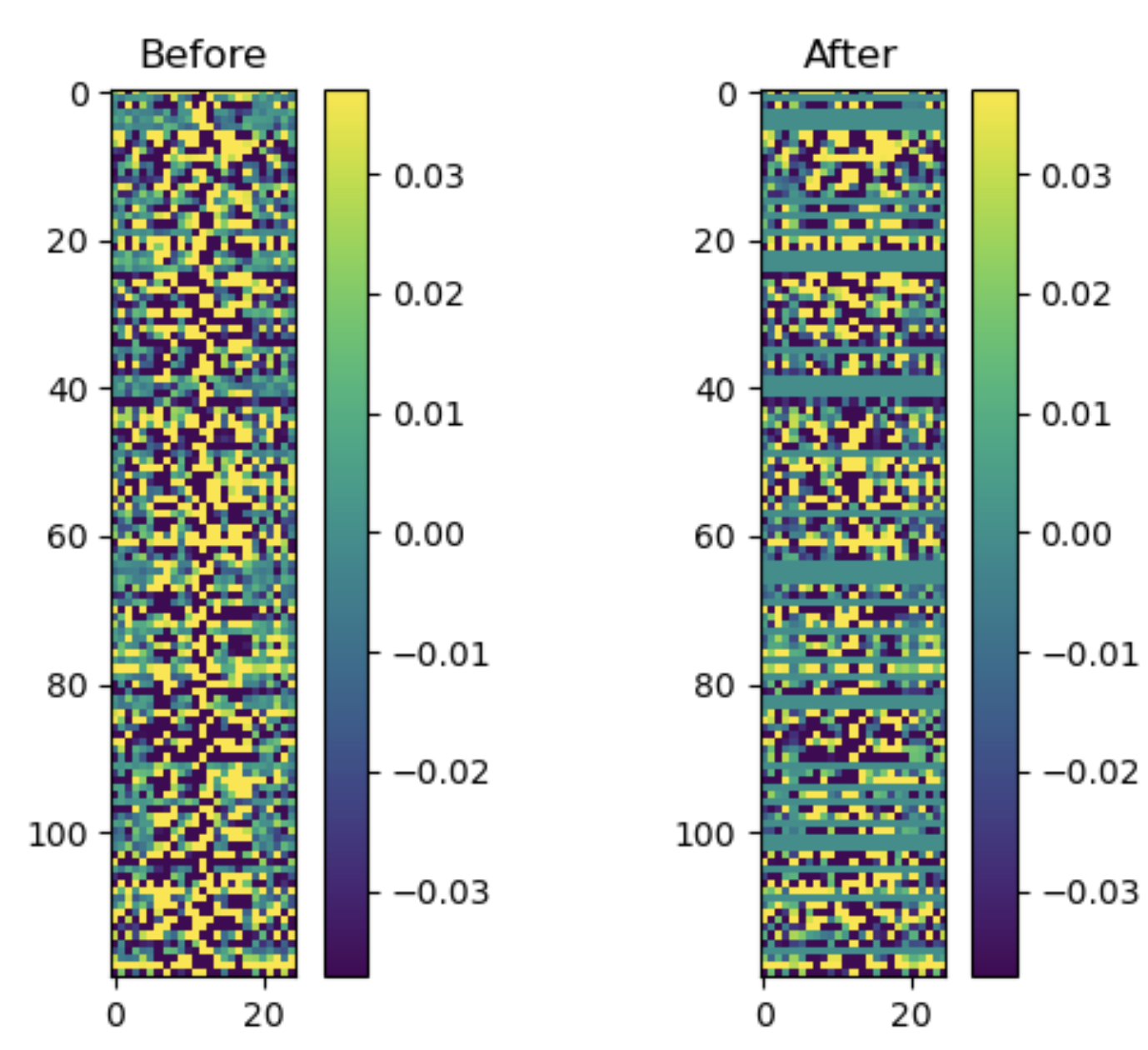}
         \caption{MobileNetv3-Small pointwise convolution before and after removing low-rank tensor factors.}
         \label{fig: mobilenetv3_pointwise_visual}
     \end{subfigure}
     \hfill
     \begin{subfigure}[b]{0.45\textwidth}
         \centering
         \includegraphics[width=\textwidth]{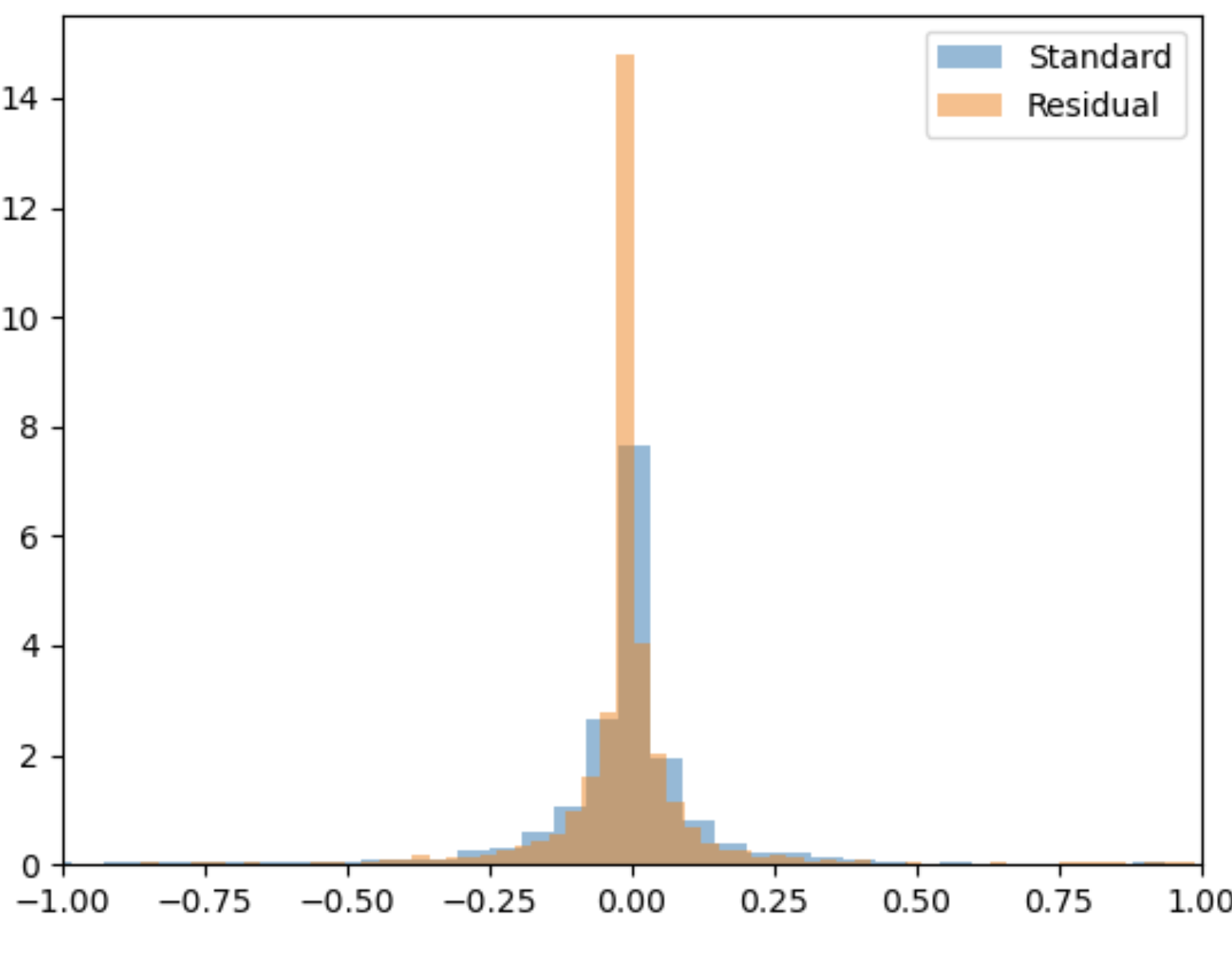}
         \caption{Effect of tensor factorization on MobileNetv3-Small pointwise convolution weights.}
         \label{fig: mobilenetv3_pointwise_hist}
     \end{subfigure}
        \caption{(a) Spatial convolutions possess regular structure that can be captured by low-rank tensor factorization. (b) Tensor factorization on spatial convolutions introduces overhead but leads to a weight distribution (in \textcolor{orange}{orange}) that is easier to prune. (c) Point-wise convolutions posses less regular structure, and tensor factorization induces fewer 0 entries. (d) Tensor factorization on pointwise convolutions introduces overhead and leads to a weight distribution (in \textcolor{orange}{orange}) that is not easier to prune.}
        \label{fig: pointwise spatial}
\end{figure}

We found that many, but not all, spatial convolutions were fit well by tensor decomposition and have low relative error. We provide an example in Figure \ref{fig: pointwise spatial} which gives representative examples of pointwise and spatial convolutions in MobileNetv3-Small. We detail the effect of our proposed compression method on the model weights. Our motivation is to ``preprocess" the weights by removing low-rank structure so that the remaining weight values are better-suited to sparse pruning. Therefore we investigate the effects of our decomposition approach on the weight distributions. We show how tensor compression can remove repeated coarse structure present in spatial convolutions, resulting in a sharply peaked weight distribution compared to the weights of the pretrained model. In contrast tensor decomposition captures less structure in the pointwise convolution and does not induce a weight distribution that is better-suited to sparse pruning. 

\subsection{Low-Rank Only}

\begin{figure}[ht]
     \centering
     \begin{subfigure}[b]{0.49\textwidth}
         \centering
         \includegraphics[width=\textwidth]{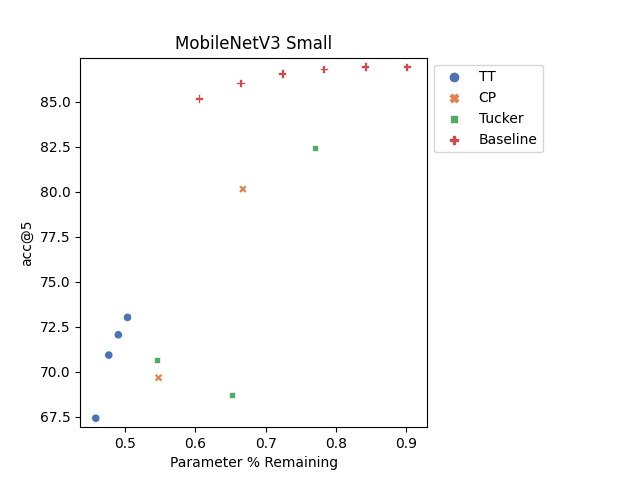}
         \caption{}
         \label{fig: mobilenet v3 small tensor only}
     \end{subfigure}
     \hfill
     \begin{subfigure}[b]{0.49\textwidth}
         \centering
         \includegraphics[width=\textwidth]{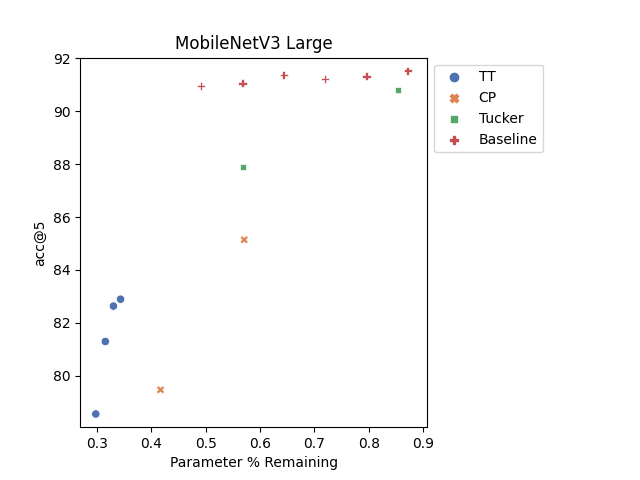}
         \caption{}
         \label{fig: mobilenet v3 large tensor only}
     \end{subfigure}
\caption{ImageNet top-5 accuracy comparison between sparse pruning and low-rank tensor decomposition on (a) MobileNet-v3-Small (b) MobileNet-v3-Large}
\label{fig: tensor only}
\end{figure}

We also considered compressing models using only post-training tensor decomposition. We take a model and perform tensor decomposition on all convolutional kernels and FC layers except the last layer. This idea was first studied in \cite{lebedev2014speeding} on larger, more compressible architectures. We report our results in Figure \ref{fig: tensor only}. We followed the same experimental setup as before, and observed that it was difficult to achieve similar accuracy with low-rank decomposition alone. We hypothesize that this is the natural result of the observation from Figure \ref{fig: layerwise residuals}. Since the model weights are not ``aproximately low-rank" it is much more challenging to decompose a model and then fine-tune. It is possible that a longer training schedule (>>50 epochs), or training from scratch in low-rank tensor format, would resolve this issue.

\FloatBarrier
\newpage

\bibliographystyle{iclr2022_conference}
\bibliography{references}

\appendix

\end{document}